\documentclass[sigconf,natbib=true,anonymous=false]{acmart}

\AtBeginDocument{%
  \providecommand\BibTeX{{%
    \normalfont B\kern-0.5em{\scshape i\kern-0.25em b}\kern-0.8em\TeX}}}

\usepackage{todonotes}
\begin{document}

\title{Risk prediction of pathological gambling on social media}
% \todo{Cooler name?}


\author{Angelina Parfenova}
\affiliation{%
  \institution{Lucerne University of Applied Sciences and Arts}
  \streetaddress{1 Th{\o}rv{\"a}ld Circle}
  \city{}
  \country{Switzerland}}
\email{angelina.parfenova@hslu.ch}

\author{Marianne Clausel}
\affiliation{%
  \institution{Université de Lorraine}
  \city{}
  \country{France}}
\email{marianne.clausel@univ-lorraine.fr}

\renewcommand{\shortauthors}{Angelina Parfenova and Marianne Clausel}

\begin{abstract}
This paper addresses the problem of risk prediction on social media data, specifically focusing on the classification of Reddit users as having a pathological gambling disorder. To tackle this problem, this paper focuses on incorporating temporal and emotional features into the model. The preprocessing phase involves dealing with the time irregularity of posts by padding sequences. Two baseline architectures are used for preliminary evaluation: BERT classifier on concatenated posts per user and GRU with LSTM on sequential data. Experimental results demonstrate that the sequential models outperform the concatenation-based model. The results of the experiments conclude that the incorporation of a time decay layer (TD) and passing the emotion classification layer (EmoBERTa) through LSTM improves the performance significantly. Experiments concluded that the addition of a self-attention layer didn't significantly improve the performance of the model, however provided easily interpretable attention scores. The developed architecture with the inclusion of EmoBERTa and TD layers achieved a high F1 score, beating existing benchmarks on pathological gambling dataset. Future work may involve the early prediction of risk factors associated with pathological gambling disorder and testing models on other datasets.  Overall, this research highlights the significance of the sequential processing of posts including temporal and emotional features to boost the predictive power, as well as adding an attention layer for interpretability. 
\end{abstract}




% %% A "teaser" image appears between the author and affiliation
% %% information and the body of the document, and typically spans the


\maketitle

\section{Introduction}
The rise of social media platforms has transformed the way people communicate, interact, and share information. With millions of users actively participating in online communities, social media platforms have become a rich source of data that can provide valuable insights into various aspects of human behavior, including mental health \cite{jiang2020detection}. In recent years, researchers have been exploring the potential of using social media data to predict and detect mental health conditions, such as depression \cite{cao2022depression}, anxiety\cite{turcan2021emotion}, and suicidal ideation \cite{allen2019convsent}.

This paper aims to contribute to the growing body of research on predicting mental health conditions on the internet. Specifically, the focus of this paper is on the prediction of pathological gambling behavior using posts from Reddit. Pathological gambling, characterized by persistent and recurrent gambling behavior that leads to significant impairment or distress, affects a significant number of individuals worldwide\cite{kim2006pathological}. By analyzing and classifying gambling behavior through social media, valuable insights can be gained to understand the prevalence and early detection of this disorder.

We begin by reviewing the studies that aimed at the prediction of pathological gambling in the context of NLP and deep learning. We critically examine existing models, particularly their handling of the online posting temporality and emotional features. In our work, we prove that the inclusion of temporal and emotional indicators can significantly improve the performance of the model. Further, we discuss the application of transformer-based models \cite{bucur2022end}, deep contextualized word embeddings \cite{campillo2022uned}, and emotion-centric approaches \cite{turcan2021emotion}, addressing challenges such as class imbalance and irregular time series. In this paper, we cover various strategies to mitigate these issues.

The methodology section outlines our approach to detecting pathological gambling using data from Reddit, detailing preprocessing, feature extraction, model development, and its evaluation. We also provide an overview of the eRisk competition \cite{parapar2021overview, parapar2022overview}, a benchmarking platform that forms the basis for our experimental validation. The experiments chapter outlines the comparison of developed models as well as how they relate to existing benchmarks on this dataset.

Overall, this paper contributes to the broader understanding of mental health prediction using machine learning. We discuss the implications of our findings, the limitations of the study, and potential directions for future research, particularly in enhancing early detection and intervention strategies for mental health conditions through social media data analysis.

\section{Related Work}
\subsection{Pathological Gambling as a condition}
Pathological gambling (PG) is a prevalent disorder characterized by persistent and recurrent maladaptive patterns of gambling behaviors \cite{miller2013interventions}. It is associated with significant personal, familial, and social costs. While currently classified as an impulse control disorder, PG shares similarities with other disorders, particularly substance addictions. The act of gambling involves placing something of value at risk in the hope of gaining something of greater value, relying on cognitive skills related to assessing risk and reward \cite{koob2010encyclopedia}.

Approximately 0.4–1.6\% of Americans meet the diagnostic criteria for PG \cite{miller2013interventions}. In most cases, gambling involves risking money, and common forms of gambling include lotteries, card games, horse and dog racing, sports betting, and slot machines. Research conducted over the past decade has provided insights into the biological features of PG and has led to the development of effective behavioral and pharmacological treatments \cite{koob2010encyclopedia}.

PG is associated with numerous negative consequences, including high rates of bankruptcy, impaired functioning, decreased quality of life, marital problems, and legal issues. If left untreated, PG tends to be a chronic and recurring condition. Individuals struggling with PG frequently experience comorbid psychiatric conditions. Common comorbid disorders include mood disorders (20–55.6\%), substance use disorders (35–76.3\%), and other impulse control disorders. Suicide attempts are also prevalent among individuals with PG, with one study reporting attempts in 17\% of the sample. Moreover, high rates of co-occurring personality disorders have been observed \cite{miller2013interventions, SOPHIA20131305}.

Overall, pathological gambling is a prevalent disorder with substantial personal, familial, and social implications. It is characterized by persistent and maladaptive patterns of gambling behaviors, leading to financial problems, impaired functioning, and diminished quality of life. Co-occurring psychiatric conditions, high rates of comorbidity, and significant gender differences further contribute to the complexity of this disorder.

\subsection{Detection of pathological gambling}

In recent years, there has been a growing interest in using natural language processing (NLP) and deep learning techniques to detect gambling disorders at an early stage using social media data. In this section, we will discuss existing approaches for pathological gambling prediction as well as the prediction of other conditions.

One of the studies that focused on the detection of PG employed an end-to-end set transformer \cite{bucur2022end}. The authors proposed the use of a set-based data-view perspective of social media posts without incorporating temporal information. Despite achieving decent results, in this paper, we argue that including temporal features in the model is crucial for better assessment. %was an for user-level classification of both depression and gambling disorder. However, they faced difficulties in incorporating the temporal component into the transformer architecture and had to resort to processing posts directly as a set, ignoring their temporal order.
In the same fashion, the study by \citep{campillo2022uned} proposed the use of TF-IDF, linguistic features, and embeddings for the concatenated set of posts. 
%They augmented their data with another Reddit dataset and trained their model using Glove embeddings and classifiers. However, they found that not all posts by users at risk contained relevant information that could be detected by an early-risk system, and labeling such posts as positive could lead to lower performance.
Sreegeethi (2022) and \cite{marmol2022sinai} used a similar approach, using embeddings for the full collection of texts for classification tasks. %Sreegeethi used random downsampling on the negative class and two transformer models, with only minor differences in performance. 
A unique approach for detecting gambling disorder was proposed by\cite{loyola2022unsl} who used a reinforcement learning-based model for decision policies. Authors developed EARLIEST, an end-to-end deep learning model that used reinforcement learning to train the model. They also used a SimpleStopCriterion decision policy to emit an alarm for a user based on the predicted class, the current delay, and the predicted positive class probability.

%\subsection{Similar tasks}

For other mental health related tasks such as detecting depression, self-harm, and suicide, recent literature shows use of the following three architectures: CNN \cite{wang2021learning, kim2020deep, lopes2021cedri}, LSTM \cite{cao2019latent, tadesse2019detection, lopes2021cedri}, and BERT-based pre-trained language models (PLMs) \cite{ragheb2021negatively, murarka2021classification, morales2021team}%, with additional complex and advanced architectures combining multiple deep learning techniques.
Besides, using the text modality from user-generated content (UGC), %used in feature representation techniques mostly involve unimodal textual datasets, with pre-trained or fine-tuned neural embeddings such as Word2Vec, fastText, GloVe, and BERT being popular choices. While most studies rely on unimodal textual datasets, 
a few research papers such as \cite{cheng2022multimodal}, have utilized multimodal architecture combining images and texts.

%The main tasks for future research are to address data imbalance and consider the temporal aspect of posts in social media to perform early detection adequately. Additionally, an explainable AI approach can improve interpretability and provide healthcare practitioners with a better understanding of the reasons behind a model's classification decisions.

\subsection{Emotion indicators for mental health}

Recent advances in mental health detection have emphasized the significance of emotional and temporal contextual cues as they are inherent to social media data. For instance, Sawhney et al. introduced STATENet \cite{sawhney2020time}, a time-aware transformer-based model, aimed at detecting suicidal risk in English tweets by analyzing the emotional spectrum of a user's historical activity. This approach highlights the potential of leveraging emotional changes over time for mental state assessment. Turcan et al.\cite{turcan2021emotion} delved into the realm of emotion detection, utilizing multi-task learning and emotion-based language model fine-tuning. Their emotion-infused models, comparable to the renowned BERT, offer a more explainable and human-like approach to psychological stress detection. The analysis of their predictive language reflects the psychological aspects of stress. Additionally, Allen et al. \cite{allen2019convsent} employed a convolutional neural network to analyze Reddit posts, incorporating the Linguistic Inquiry and Word Count (LIWC) dictionary. Their methodology, which examines topics, emotional experiences, and stylistic elements, underscores the efficacy of linguistic analysis in mental health prediction.

Thus, emotional and temporal contextual cues have proven to be valuable indicators for detecting mental health issues such as suicidal ideation and psychological stress in social media posts. These cues have been leveraged through various NLP techniques, including transformer-based models, deep contextualized word representations, and emotion-infused models. The use of these techniques has resulted in significant improvements in mental health detection, making it possible to identify individuals at risk and provide timely interventions.

% \subsection{Scope of Interpretability}
Still, a significant challenge in building ML-based mental health assessment systems is the lack of a feedback loop to capture real-world outcomes. %It is difficult to determine if a user has committed suicide, how many incidents of self-harm have occurred, or how long a user has been suffering from depression. 
%Therefore, there is a need for explainable AI (XAI) techniques in high-risk domains such as healthcare to enable healthcare practitioners to understand why an AI model classified a user as positive or negative, along with the associated symptoms and causes.
Some studies have attempted to extract this information using XAI techniques such as attention weight analysis, deep attention network-based clustering, and understanding its hidden layers' activation patterns \cite{zogan2022explainable, uban2021explainability, farruque2021explainable, naseem2022early}. In the next section, we describe the main challenges in mental health prediction as well as some approaches to overcome them.
\subsection{Main challenges}
\subsubsection{Class Imbalance} A prominent issue in mental health data analysis is the class imbalance problem, wherein datasets typically feature a lower number of depressed individuals compared to non-depressed ones. This imbalance presents a significant obstacle to correct predictions\cite{parapar2021overview}.  In this chapter, we discuss two studies that propose solutions to handle imbalanced datasets for depression detection using text data. 

\cite{cong2018xa} proposed a model to detect users with depression specifically in imbalanced datasets using text data. They used a distributed neural word embedding created by lookups in a domain-specific embedding matrix. Their model was a cascaded X-A-BiLSTM model with two components - XGBoost and Bi-LSTM with Attention mechanism. The XGBoost component was used to handle data imbalance, followed by the Bi-LSTM with an Attention mechanism for improved classification accuracy. \cite{rao2020knowledge} also proposed a model to detect users with depression in imbalanced datasets using text data. They infused knowledge triples into BERT neural word embeddings (KFB) and used a BiGRU with Attention mechanism (KFB-BiGRU-Att) as the primary model. 

Apart from using special machine learning techniques to cope with imbalanced datasets, one can use augmentation techniques. For instance, \cite{wu2020deep} added external public heterogeneous information such as traffic, weather, environment, population, and living conditions. They merged the feature representation vectors from both sets and used them to train a deep neural network for depression detection.

In another study, authors scraped additional data from Reddit to address the imbalanced dataset problem in depression detection\cite{campillo2022uned}. The authors used the PRAW Python Reddit API Wrapper to extract new data from different subreddits related to mental health and depression. They manually reviewed the results to ensure that the users had been officially diagnosed with clinical depression and had a sufficient number of submissions. In contrast, in another paper authors used downsampling to cope with the imbalanced data problem in the early detection of depression\cite{parapar2021overview}. They randomly removed a substantial amount of data from the negative class to balance the dataset. 

\subsubsection{Time irregularity} Another inherent characteristic of social network data is the irregularity in the posting time. The irregularity and inconsistency of time intervals between posts can lead to uncertainty in identifying and understanding the mental state of users. Several research papers have proposed novel approaches to handle such irregularities, and this chapter will discuss some of them.

For example, \cite{sawhney2020time} proposed a time-aware LSTM that augments linguistic models with historical context to identify suicidal intent in English tweets. The model utilizes temporal contextual cues of a user's historical activity on social media to assess their suicide risk. The same architecture was used by \cite{cheng2022multimodal} where they use hashtags to label users. Text, image, and posting time are jointly used to detect depression via concatenating them as features for the network. 

There are other approaches such as using GRU neural network, which in essence is a simpler version of LSTM that is easy and efficient to train \cite{chung2014empirical}. Inspired by this approach \cite{cao2022depression} proposes a model for depression prediction based on BiAttention-GRU by analyzing text, speech, and facial expression features associated with depression. GRU is used to extract temporal information, and the Softmax layer is used for final prediction. Another approach was taken recently by \cite{bucur2023s} where authors proposed a flexible time-enriched multimodal transformer architecture for detecting depression from social media posts. The authors use pre-trained models to extract image and text embeddings and enrich their model with the relative time between posts using time2vec positional embeddings. 

\subsubsection{Interpretability}

In the domain of predicting diagnoses from user posts on the Internet, it becomes crucial to understand what the model is focusing on and which parts of the text contribute to its decision-making process. This is especially important in the context of human psychology, where the nuances and subtleties of language play a significant role. While attention mechanisms provide a means to capture relevant information, interpreting the attention weights can offer valuable insights into the model's reasoning process, making it more transparent.

A noteworthy study by \cite{cheng2022multimodal} explores the application of attention weights in the context of mental health diagnosis prediction. The author demonstrates the attention weights assigned to each post, allowing researchers and practitioners to observe which specific parts of the text the model is paying attention to in order to generate a general conclusion on the user level. This approach not only enhances the interpretability of the model but also provides valuable insights into the features or themes that contribute to the diagnosis prediction.

In this paper, we aim to incorporate an attention mechanism to enhance the interpretability of our model as well. By analyzing the attention weights, we can gain a deeper understanding of how the model processes and weighs different parts of the input text. However, it remains to be seen whether the addition of the attention measure will improve the overall quality and performance of the model. Nevertheless, the inclusion of interpretability features is a crucial step towards building trust and understanding in the predictions made by the model, especially in sensitive domains such as mental health assessment.

\subsection{Dataset}

\begin{figure}[h]
  \centering
  \includegraphics[width=0.45\textwidth]{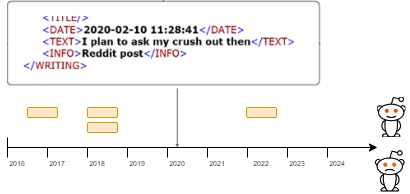}
  \caption{Illustration of the data}
  \label{fig:data}
\end{figure}

The dataset on pathological gambling was provided during an annual eRisk competition \cite{parapar2022overview}. Participants are provided with a dataset containing social media posts from users who have been diagnosed with some mental health disorder and those who have not (Fig.\ref{fig:data}). The objective of the competition is to develop machine learning models that can accurately classify users into the appropriate groups.

Examining the data on pathological gambling (Table \ref{tab:dataset}), we can tell that it contains a significant amount of text, with a wide range of post counts per instance. However, there is a severe imbalance in the labeling of instances, with only 245 instances being labeled as positive, and 4139 instances being labeled as negative. This skewness in labeling can have a significant impact on the performance of any machine learning models trained on this dataset. Therefore, it is crucial to address this imbalance issue before proceeding with any further research.

\begin{table}
    \centering
    \caption{Dataset Statistics}
    \begin{tabular}{p{5cm}l}
    \toprule
    \#users &  4384 \\
    \#Positive labels & 245 \\
      \#Negative labels & 4139 \\
      \texttt{min}(\#post per user) & 3\\
      \texttt{max}(\#post per user)  & 2002 \\
      \texttt{avg}(\#post per user) & $520_{551}$ \\    
      \texttt{min}(text-length per post) & 0 \\
      \texttt{max}(text-length per post) &  10157 \\
      \texttt{avg}(text-length per post) & $28_{81}$ \\
    
    \bottomrule
    \end{tabular}
    
    \footnotesize{Note: Subscript refers to standard deviation; conc. refers to concatenated}
    \label{tab:dataset}
\end{table}

\subsection{Problem Formulation}
We define a sequence of social media posts as $S = \{(m_i,t_i): i=1 \text{ to } |S|\}$, where $m_i$ is the $i^{th}$ post by a user on timestamp $t_i$, and $y$ denotes the binary label of the user's mental health status (e.g., \textit{pathological gambling}).

\subsection{Preprocessing}
\subsubsection{Tackling imbalance}
To address data imbalance, random downsampling is applied during the preprocessing phase. The original dataset, $D$, consisting of user post sequences and binary labels, is defined as:
\[ D = \{ (S_i, y_i) : S_i = \{(m_{ij}, t_{ij}) : j = 1 \text{ to } |S_i|\}, y_i \in \{0,1\}, i = 1 \text{ to } N \} \]
where $N$ is the total number of users.

Assuming $N_0 > N_1$ for the majority ($y=0$) and minority ($y=1$) classes respectively, random downsampling selects a subset from the majority class at random to match the minority class size, resulting in a new balanced dataset $D'$:
\[ D' = D_{\text{minority}} \cup \text{RandomSubset}(D_{\text{majority}}, N_1) \]
where $\text{RandomSubset}(D_{\text{majority}}, N_1)$ is the subset of $D_{\text{majority}}$ with size $N_1$. The balanced dataset is then used for training classification models, providing equal representation for both classes.

\subsubsection{Temporal irregularity}

\begin{figure}
  \centering
  \includegraphics[width=0.39\textwidth]{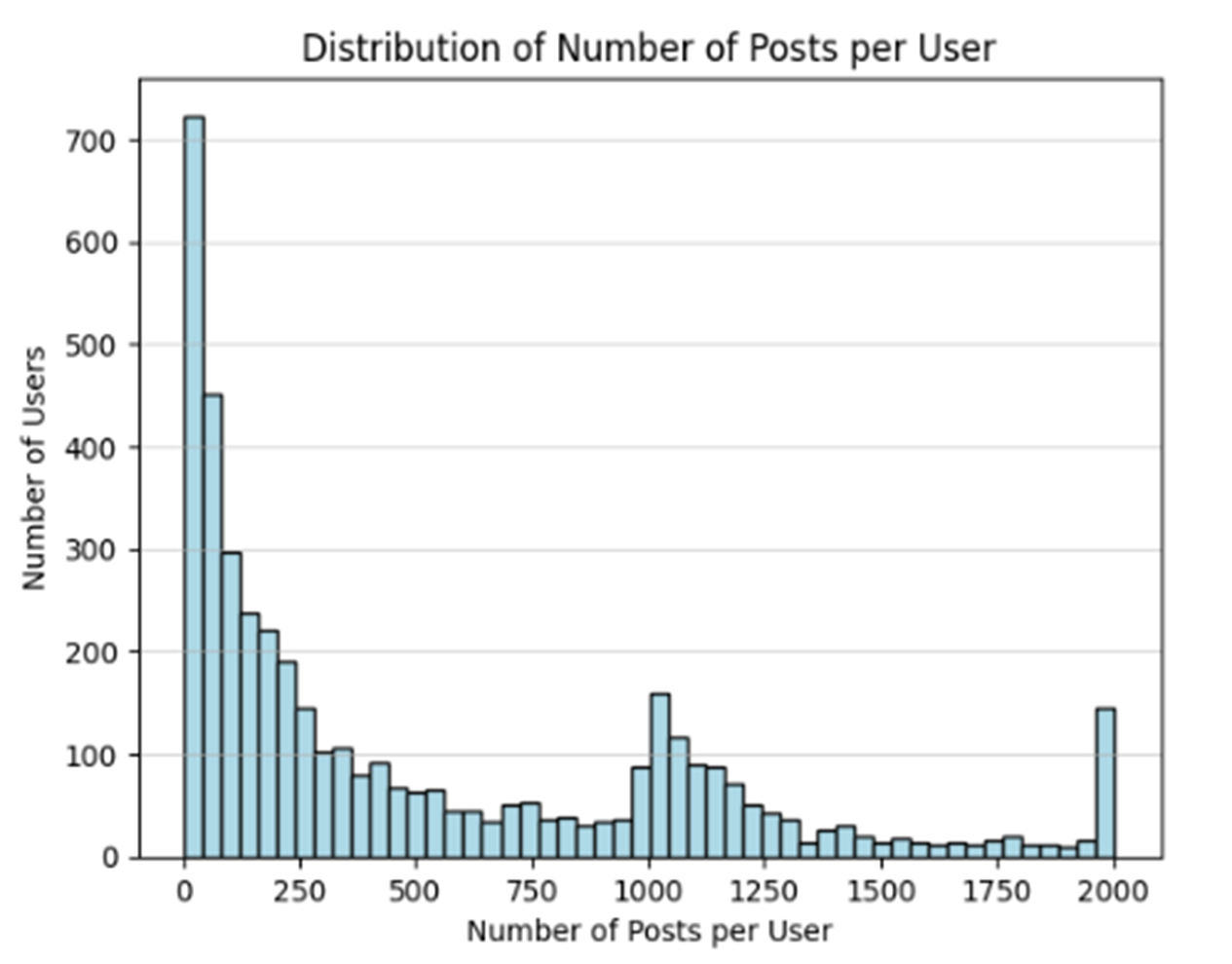}
  \caption{The number of posts per user, demonstrating the variation addressed by sequence padding}
  \label{fig:distribution}
\end{figure}

Fig.\ref{fig:distribution} illustrates a notable variation in the number of posts across the dataset. To address this issue, the sequence padding was used in the following fashion:

Consider a set of sequences $S = \{s_1, s_2, \ldots, s_m\}$, where each sequence $s_i$ is a vector representing posts by a user, and $m$ denotes the total number of such sequences. The length of an individual sequence $s_i$ is denoted by $|s_i|$. The maximal length of a sequence within our dataset is given by $L_{\text{max}} = \max_{i}(|s_i|)$.

The padding transformation is applied to each sequence $s_i$ to produce a new sequence $s_i'$, where:

\begin{equation*}
s_i' = 
\begin{cases} 
s_i & \text{if } |s_i| = L_{\text{max}} \\
(s_i, 0, \ldots, 0) & \text{if } |s_i| < L_{\text{max}}
\end{cases}
\end{equation*}

In this operation, $0$ denotes the padding element used to fill the sequence $s_i$ to reach the desired length $L_{\text{max}}$. This ensures uniformity across all sequences, facilitating their processing by the model. By employing the longest sequence as the standard length, $L_{\text{max}}$, the model is provided with the most extensive possible view of the data, ensuring that no valuable information is omitted during processing.

\section{Experimentation}

\subsection{Baselines}

In this section, we explore two baseline models designed for analyzing user-generated text data. 

\paragraph{Text-Baseline} This model is designed to process entire blocks of text submitted by users. Its primary function is to understand and predict user behavior or characteristics based on their textual input. In our case, the model takes as input all posts by a single user concatenated $||_{i=1}^{|S|} m_i$ to \texttt{BERT} based pretrained language model and uses the \texttt{[CLS]} embedding for predicting $\hat{y}$.

\paragraph{Seq-Baseline} The Seq-Baseline model takes a slightly different approach, focusing on the sequential nature of text. It aims to capture the progression or changes in user behavior over time. This model takes as input the sequence of posts into account omitting the timestamp information. It encodes the text using \texttt{BERT}-based PLM and then uses \texttt{GRU} and \texttt{LSTM} model to capture the temporal progression of the user's pathological gambling state to predict $\hat{y}$. 

Both models aim to provide a comprehensive understanding of user behavior through textual analysis and are widely used in previous works. However, in the current research, we want to improve the overall quality of the model by considering new features: emotional indicators of PG and decay of post importance in final class prediction.

\subsection{Proposed model}

\begin{figure}
  \centering
  \includegraphics[width=0.45\textwidth]{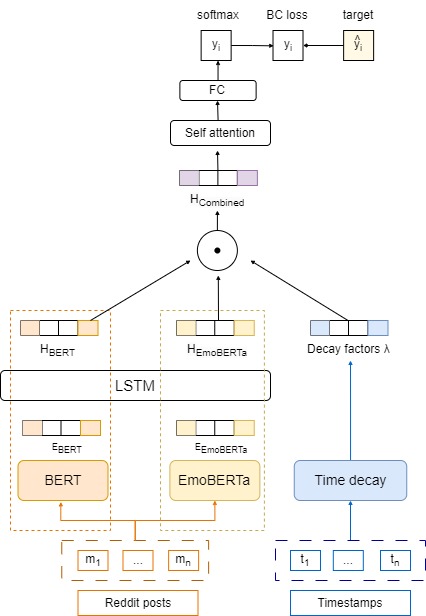} 
  \caption{Model architecture}
  \label{fig:diag}
\end{figure}

The architecture, as depicted in Figure \ref{fig:diag}, is designed to process sequences of textual data paired with timestamps for binary classification in mental health detection.

Let \( S = \{m_1, m_2, \ldots, m_n\} \) be the sequence of posts extracted from a social media user, where \( n \) denotes the sequence length, and \( T = \{t_1, t_2, \ldots, t_n\} \) be the corresponding timestamps. The model includes the following components:

\begin{itemize}
    \item \textbf{Embedding Layers:} BERT and EmoBERTa classifiers independently encode the sequence of posts \( S \) into two distinct dense representations, \( E_{\text{BERT}} \) and \( E_{\text{EmoBERTa}} \), respectively. A time decay layer computes decay factors \( D \) for the timestamps \( T \).
    \begin{align*}
        E_{\text{BERT}} &= \text{BERT}(S), \\
        E_{\text{EmoBERTa}} &= \text{EmoBERTa}(S), \\
        \lambda &= \text{TimeDecayLayer}(T), \\
        \text{TimeDecayLayer}(t_i) &= e^{-\frac{t_i - \text{roll}(t_i, 1)}{86400}} \quad \forall t_i \in T.
    \end{align*}

\item \textbf{LSTM Layers:} Two separate LSTM layers process \( E_{\text{BERT}} \) and \( E_{\text{EmoBERTa}} \) to produce corresponding series of hidden states \( H_{\text{BERT}} \) and \( H_{\text{EmoBERTa}} \).

\item \textbf{Multiplication and Time Decay Integration:} The hidden states from both LSTM layers are element-wise multiplied to form a single combined hidden state sequence \( H_{\text{combined}} = \{H_1, H_2, \ldots, H_n\} \), which is then element-wise multiplied with the time decay factors $\lambda$.
\begin{align*}
    H &= H_{\text{BERT}} \odot H_{\text{EmoBERTa}} 
\end{align*}
\begin{align*}
     H_{\text{combined}} &= H \odot \lambda.
\end{align*}

\item \textbf{Attention Mechanism:} Assigns weights to the elements in \( H_{\text{combined}} \) based on their relevance, resulting in an attended output \( \text{att\_out} \) and attention weights.
\begin{align*}
    (\text{att\_out}, \text{attention\_weights}) &= \text{Attention}(H_{\text{combined}}).
\end{align*}

    \item \textbf{Fully Connected and Softmax Layers:} The attended output is then passed through a fully connected layer and softmax activation to predict the probability distribution over two classes.
    \begin{align*}
        Y &= \text{Softmax}(\text{FC}(\text{att\_out})).
    \end{align*}
    
    \item \textbf{Loss Function:} The binary cross-entropy loss function is utilized to optimize the model by minimizing the difference between predicted probabilities $Y$ and the target labels.
    \begin{align*}
        L &= \text{BC\_loss}(Y, \text{target}).
    \end{align*}
\end{itemize}

The model's parameters are optimized via the Adam optimizer, with a learning rate scheduled according to the epoch number. Post-training, the model is evaluated using various performance metrics on the test data.

\subsection{Model Training and Parameters}

The training of the proposed model was conducted on a high-performance computing system equipped with a NVIDIA Tesla V100 SXM2 GPU, optimized for deep learning tasks. Training parameters included 10 epochs, with each epoch entailing multiple iterations of parameter updates based on loss calculations. A batch size of 32 was chosen to balance training speed and memory usage, with an initial learning rate set at 0.001. The Adam optimizer was utilized for its adaptability and efficiency in parameter updates, and the binary cross-entropy loss function guided the optimization for the binary classification task. Additional strategies to enhance model performance included model checkpointing, saving the model with the lowest validation loss for later use. 

\section{Experimental results}

Below is a table presenting the results of the experiments conducted. Each model was run 10 times with different seeds and the average of performance metrics on the test set was taken. Initially, a GRU+Dropout (GRUd) architecture was utilized without incorporating temporal or emotive components. The results indicate that the sequential model achieved a relatively high F1 score compared to the model that relied primarily on concatenated texts, which was the primary approach for this dataset.

\subsection{Implemented models}

Overall there were several implemented models before the aforementioned one. Here is the table with their properties (Table \ref{tab:models}. In the experiments section their performance will be covered.

\begin{table}[htbp]
  \centering
  \caption{Models with Different Features}
  \begin{tabular}{|c|c|c|c|}
    \hline
    Model  & Time decay & Emotions & Attention \\
    \hline
    GRUd   & & & \\
    \hline
    GRUdTd  & $\checkmark$ & & \\
    \hline
    LSTMTd  & $\checkmark$ & & \\
    \hline
    LSTMTdA & $\checkmark$ & & $\checkmark$ \\
    \hline
    EmoLSTMTd & $\checkmark$ & $\checkmark$ & \\
    \hline
    EmoLSTMTdA & $\checkmark$ & $\checkmark$ & $\checkmark$ \\
    \hline
  \end{tabular}
  \label{tab:models}
\end{table}

It is worth noting that the model using concatenated vectors from Emotion classification results and BERT embeddings performed significantly worse compared to the model where Emotion classification results were passed through the network layer and then multiplied with the sentence embeddings layer. The concatenated vectors approach yielded a score that is close to random performance. This observation emphasizes the importance of incorporating the emotion classification results appropriately to enhance the model's performance. To explore the significance of time, the inclusion of time decay was incorporated, as described earlier. Notably, the results demonstrated a significantly higher F1 score on the test sample, indicating that the inclusion of time decay in the model enhances its predictive power (Table \ref{tab:results}).

Additionally, the LSTM model was employed as an experiment, which yielded considerably better results on this dataset compared to the GRU model. As a result, it was decided to employ the LSTM model for all subsequent experiments, including the addition of emotional embedding and attention layers for the XAI (Explainable Artificial Intelligence) component.

\begin{table*}[h]
\centering
\caption{Results of Model Comparison}
\begin{tabular}{lccccccc}
\hline
Model & Accuracy & Precision & Recall & F1 & \multicolumn{1}{c}{Std} & AUROC & AUPRC \\
\hline
BERT classifier & 0.827 & 0.763 & 0.966 & 0.836 & ±0.02 & 0.924 & 0.922 \\
\hline
GRUd & 0.911 & 0.905 & 0.916 & 0.908** & ±0.007 & 0.895 & 0.87 \\
GRUdTd & 0.954 & 0.932 & 0.918 & 0.929** & ±0.05 & 0.956 & 0.94 \\
\hline
LSTMTd & 0.938 & 0.941 & 0.953 & 0.946* & ±0.03 & 0.935 & 0.914 \\
LSTMTdA & 0.963 & 0.934 & 0.958 & 0.948 & ±0.01 & 0.935 & 0.915 \\
\hline
LSTMTd & 0.938 & 0.941 & 0.953 & 0.946 & ±0.03 & 0.935 & 0.914 \\
EmoLSTMTd & 0.955 & 0.958 & 0.951 & 0.954* & ±0.01 & 0.951 & 0.934 \\
EmoLSTMTdA & 0.952 & 0.955 & 0.951 & 0.955 & ±0.02 & 0.949 & 0.925\\
\hline
\end{tabular}

\footnotesize{Note: * Results are significantly better than the previous row in the table at the 0.005 level, ** 0.001 level (p-value) according to the Wilcoxon test.}
\label{tab:results}
\end{table*}

The results (Table \ref{tab:results}) indicate that adding EmoBERTa layer yields higher performance. This can be explained by the fact that When applying an LSTM layer to the emotion input, the model captures the temporal dependencies or patterns within this information. This is particularly beneficial when the sentiment information exhibits sequential or temporal characteristics, where emotions evolve or change over time within the input sequence. LSTM layer captures the dependencies and patterns in the emotion data enabling the model to consider the past context of emotions and utilize it for making predictions at each time step.

As a subsequent experiment, a self-attention layer was added to the model, anticipating that it would improve performance like in related work \cite{cheng2022multimodal}. However, the model did not exhibit any significant change in its predictive power compared to the model without it. Nonetheless, the interpretation of attention scores remains an essential component of Explainable AI. Therefore, the decision was made to train the model and observe its attention patterns to gain insights into its decision-making process.

The Wilcoxon test was used in this study to assess the statistical significance of the performance differences between the models. The test compares the paired observations of the models' performance metrics to determine if there is a significant improvement or difference between them. In our study this test was employed to compare the F1 scores of different models (Table \ref{tab:wilcoxon}). The test results are denoted in the table as asterisks (*) and (**) indicating the significance level (0.005 and 0.001) at which the results are better than the previous model.

For example, in the row comparing the "LSTMTd" model to the "GRUdTd" model, an asterisk (*) is present in the "F1" column, indicating that the F1 score of the "LSTMTd" model is significantly better than that of the "GRUd" model. Similarly, in the row comparing the "EmoLSTMTd" model to the "LSTMTd" model, an asterisk (*) is present in the "F1" column, indicating that the F1 score of the "EmoLSTMTd" model is significantly better than that of the "LSTMTd" model.

\begin{table}[h]
\centering
\caption{Wilcoxon Test Results}
\begin{tabular}{llll}
\hline
Model Comparison & Z-value & p-value & Sig \\
\hline
BERTClassifier vs GRUd & -3.0594 & 0.00222 & Yes \\
GRUd vs GRUdTd & -2.6502 & 0.00804 & Yes \\
GRUdTd vs LSTMtd & -2.1405 & 0.03236 & Yes \\
LSTMtd vs LSTMtdA & -0.6668 & 0.50286 & No \\
LSTMtd vs EmoLSTMtd & -2.0894 & 0.03662 & Yes \\
EmoLSTMtd vs EmoLSTMtdA & -0.489 & 0.62414 & No \\
\hline
\end{tabular}
\label{tab:wilcoxon}
\end{table}

\subsubsection{Emotional input}

In our case, we used EmoBERTa classification results as the input to LSTM. To obtain these results, we loaded a pretrained EmoBERTa model and utilized it to predict seven types of emotions for each post. This classifier developed by \cite{kim2021emoberta}, provides numerous advantages. These include an enhanced understanding of the text, improved performance, and the ability to detect nuanced emotions. 

In a related study conducted by \cite{bucur2023s}, incorporating EmoBERTa prediction results into a classification task yielded relatively higher model performance. Our study also confirms that the inclusion of this information significantly boosts the performance of a model that is aimed at predicting pathological gambling (Table \ref{tab:results}).

\subsubsection{Attention module}

The attention mechanism, essential in modern natural language processing enables models like the transformer to selectively focus on relevant parts of input data \cite{vaswani2017attention}. This mechanism assigns weights to different segments of data, assessing their importance through a comparison of query, key, and value elements. By applying a softmax function to these comparisons, the model generates a weighted representation that captures crucial information across various data points. 

\begin{figure*}
  \centering
  \includegraphics[width=0.7\textwidth]{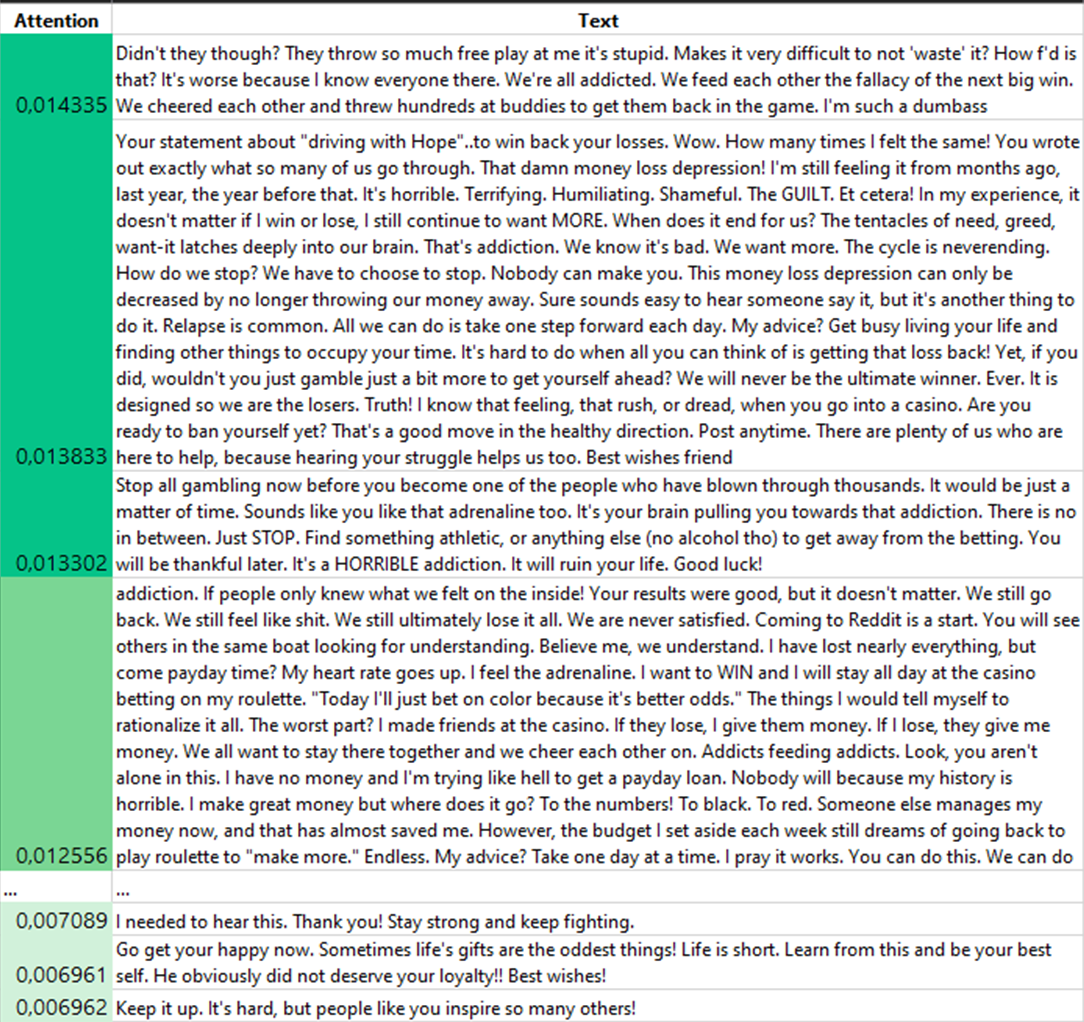}
  \caption{Attention weights for several observations}
  \label{fig:attention}
\end{figure*}

In the context of our experiments, adding an attention layer to the model did not significantly improve its predictive power. However, it was still worth looking at the attention scores to see which posts contributed to the positive classification. Below are the attention weights for several observations, sorted in descending order. The highest weights were for rather long posts, with the use of caps lock, exclamation marks, and emotional coloring such as guilt and shame, which could be a sign of pathological gambling (Fig. \ref{fig:attention}).

\subsubsection{Comparison with other benchmarks}

Let's take a look at the performance of the model in comparison to previous models tried on this dataset (Table \ref{tab:benchmarks}).

\begin{table}[h]
\centering
\caption {Models for this dataset \cite{parapar2022overview}}
\begin{tabular}{lccc}
\hline
Model & Precision & Recall & F1 \\
\hline
UNED-NLP (ANN) & 0.809 & 0.938 & 0.869 \\
SINAI (Roberta large) & 0.908 & 0.728 & 0.808 \\
BLUE (Set Transformer) & 0.052 & \textbf{1} & 0.099 \\
UNSL (RL) & 0.052 & 0.988 & 0.1 \\
IISERB (BERT classifier) & \textbf{1} & 0.074 & 0.138 \\
\hline
EmoLSTMTd & 0.958 & 0.951 & \textbf{0.954}\\
\hline

\end{tabular}
\label{tab:benchmarks}
\end{table}

Our model outperforms the previous models in the competition, as evidenced by its higher F1-score. However, it's important to acknowledge the limitations of the dataset used in our study. The dataset is relatively small, consisting of only 490 objects with a balanced sample of two classes. To fully evaluate the predictive power of our model, it should be tested on larger and more diverse datasets.

For instance, when considering self-harm data, the available dataset contains only around 90 objects \cite{parapar2022overview}, which is insufficient to thoroughly assess the performance of the same model. Furthermore, different types of psychological problems may not exhibit the same time or emotion dependencies as the dataset used in this study. Therefore, other datasets may require entirely different models or approaches to achieve accurate predictions.

In summary, while our model demonstrates superior performance on the current dataset, further research and testing on larger and more varied datasets are necessary to validate its effectiveness across different psychological domains and ensure its generalizability.

\section{Conclusion}
In this section, we outline the main contributions of this paper as well as its limitations. 

The conducted experiments proved that using sequential models is the best fit for the analysis of social media data. Processing posts directly as a set proved to have significantly lower performance even while using advanced architectures such as transformers. The inclusion of the time decay layer demonstrated a significant increase in F1 score, proving that in mental health prediction later posts bear more weight in making diagnosis. This approach suggests that social media posts may mirror the nature of patient diagnoses, where recent information is more important than older data. 

During the experiments, it was discovered that the inclusion of emotion classification results by passing them through the LSTM layer significantly improves the performance of the model. It shows that emotions serve as potential indicators of pathological gambling and may change over time which is captured by sequential post processing. Furthermore, inclusion of self attention didn't significantly increase the F1 score, however provided the possibility of interpreting results. By looking at attention scores it is easier to detect which posts contributed to the final prediction.

Although our model's F1-score of 0.95 is the highest reported on the available benchmarks for this dataset, it is important to note that this architecture is not a groundbreaking advancement in Natural Language Processing. The MTAN architecture with an attention mechanism incorporated into LSTM, as demonstrated in the work by \cite{cheng2022multimodal}, achieved similar results on another dataset. 

Another limitation of our study is the scarcity of annotated data in terms of time. Some posts had a gap of several years which could contribute to the lower predictive power. Research into time-aware architectures remains a relatively unexplored area. This issue was also complicated by the scarcity and imbalance of the dataset itself. Random downsampling significantly reduced the available data, which may have negatively contributed to the predictions,

It is important to acknowledge that our study did not specifically focus on early risk detection, which involves implementing a stopping mechanism to predict mental health conditions as early as possible based on users' posts. Early detection plays a crucial role in preventing negative outcomes and facilitating early-stage diagnosis of patients and addressing this aspect falls within the scope of future work.

Future research will focus on developing an early risk prediction system that utilizes the proposed model to assess the likelihood of mental health issues. Further improvements to the model can also be made especially with larger datasets and testing the suggested architecture for prediction of other mental health problems. Further work will also include refining predictions using other XAI methods. In such a sensitive domain, reliance on 'black-box' models is not viable; understanding the underlying decision-making processes is crucial for responsible and trustworthy applications in mental health assessments.

\bibliographystyle{ACM-Reference-Format}
\bibliography{sample-base}


\begin{thebibliography}{35}

%%% the Web address.

\ifx \showCODEN    \undefined \def \showCODEN     #1{\unskip}     \fi
\ifx \showDOI      \undefined \def \showDOI       #1{#1}\fi
\ifx \showISBNx    \undefined \def \showISBNx     #1{\unskip}     \fi
\ifx \showISBNxiii \undefined \def \showISBNxiii  #1{\unskip}     \fi
\ifx \showISSN     \undefined \def \showISSN      #1{\unskip}     \fi
\ifx \showLCCN     \undefined \def \showLCCN      #1{\unskip}     \fi
\ifx \shownote     \undefined \def \shownote      #1{#1}          \fi
\ifx \showarticletitle \undefined \def \showarticletitle #1{#1}   \fi
\ifx \showURL      \undefined \def \showURL       {\relax}        \fi
% invisible to TeX
\providecommand\bibfield[2]{#2}
\providecommand\bibinfo[2]{#2}
\providecommand\natexlab[1]{#1}
\providecommand\showeprint[2][]{arXiv:#2}

\bibitem[Allen et~al\mbox{.}(2019)]%
        {allen2019convsent}
\bibfield{author}{\bibinfo{person}{Kristen Allen}, \bibinfo{person}{Shrey Bagroy}, \bibinfo{person}{Alex Davis}, {and} \bibinfo{person}{Tamar Krishnamurti}.} \bibinfo{year}{2019}\natexlab{}.
\newblock \showarticletitle{ConvSent at CLPsych 2019 Task A: using post-level sentiment features for suicide risk prediction on Reddit}. In \bibinfo{booktitle}{\emph{Proceedings of the Sixth Workshop on Computational Linguistics and Clinical Psychology}}. \bibinfo{pages}{182--187}.
\newblock


\bibitem[Bucur et~al\mbox{.}(2022)]%
        {bucur2022end}
\bibfield{author}{\bibinfo{person}{Ana-Maria Bucur}, \bibinfo{person}{Adrian Cosma}, \bibinfo{person}{Liviu~P Dinu}, {and} \bibinfo{person}{Paolo Rosso}.} \bibinfo{year}{2022}\natexlab{}.
\newblock \showarticletitle{An end-to-end set transformer for user-level classification of depression and gambling disorder}.
\newblock \bibinfo{journal}{\emph{arXiv preprint arXiv:2207.00753}} (\bibinfo{year}{2022}).
\newblock


\bibitem[Bucur et~al\mbox{.}(2023)]%
        {bucur2023s}
\bibfield{author}{\bibinfo{person}{Ana-Maria Bucur}, \bibinfo{person}{Adrian Cosma}, \bibinfo{person}{Paolo Rosso}, {and} \bibinfo{person}{Liviu~P Dinu}.} \bibinfo{year}{2023}\natexlab{}.
\newblock \showarticletitle{It's Just a Matter of Time: Detecting Depression with Time-Enriched Multimodal Transformers}.
\newblock \bibinfo{journal}{\emph{arXiv preprint arXiv:2301.05453}} (\bibinfo{year}{2023}).
\newblock


\bibitem[Campillo-Ageitos et~al\mbox{.}(2022)]%
        {campillo2022uned}
\bibfield{author}{\bibinfo{person}{Elena Campillo-Ageitos}, \bibinfo{person}{Juan Martinez-Romo}, {and} \bibinfo{person}{Lourdes Araujo}.} \bibinfo{year}{2022}\natexlab{}.
\newblock \showarticletitle{UNED-MED at eRisk 2022: depression detection with TF-IDF, linguistic features and Embeddings}.
\newblock \bibinfo{journal}{\emph{Working Notes of CLEF}} (\bibinfo{year}{2022}), \bibinfo{pages}{5--8}.
\newblock


\bibitem[Cao et~al\mbox{.}(2019)]%
        {cao2019latent}
\bibfield{author}{\bibinfo{person}{Lei Cao}, \bibinfo{person}{Huijun Zhang}, \bibinfo{person}{Ling Feng}, \bibinfo{person}{Zihan Wei}, \bibinfo{person}{Xin Wang}, \bibinfo{person}{Ningyun Li}, {and} \bibinfo{person}{Xiaohao He}.} \bibinfo{year}{2019}\natexlab{}.
\newblock \showarticletitle{Latent suicide risk detection on microblog via suicide-oriented word embeddings and layered attention}.
\newblock \bibinfo{journal}{\emph{arXiv preprint arXiv:1910.12038}} (\bibinfo{year}{2019}).
\newblock


\bibitem[Cao et~al\mbox{.}(2022)]%
        {cao2022depression}
\bibfield{author}{\bibinfo{person}{Yongzhong Cao}, \bibinfo{person}{Yameng Hao}, \bibinfo{person}{Bin Li}, {and} \bibinfo{person}{Jie Xue}.} \bibinfo{year}{2022}\natexlab{}.
\newblock \showarticletitle{Depression prediction based on BiAttention-GRU}.
\newblock \bibinfo{journal}{\emph{Journal of Ambient Intelligence and Humanized Computing}} \bibinfo{volume}{13}, \bibinfo{number}{11} (\bibinfo{year}{2022}), \bibinfo{pages}{5269--5277}.
\newblock


\bibitem[Cheng and Chen(2022)]%
        {cheng2022multimodal}
\bibfield{author}{\bibinfo{person}{Ju~Chun Cheng} {and} \bibinfo{person}{Arbee~LP Chen}.} \bibinfo{year}{2022}\natexlab{}.
\newblock \showarticletitle{Multimodal time-aware attention networks for depression detection}.
\newblock \bibinfo{journal}{\emph{Journal of Intelligent Information Systems}} \bibinfo{volume}{59}, \bibinfo{number}{2} (\bibinfo{year}{2022}), \bibinfo{pages}{319--339}.
\newblock


\bibitem[Chung et~al\mbox{.}(2014)]%
        {chung2014empirical}
\bibfield{author}{\bibinfo{person}{Junyoung Chung}, \bibinfo{person}{Caglar Gulcehre}, \bibinfo{person}{KyungHyun Cho}, {and} \bibinfo{person}{Yoshua Bengio}.} \bibinfo{year}{2014}\natexlab{}.
\newblock \showarticletitle{Empirical evaluation of gated recurrent neural networks on sequence modeling}.
\newblock \bibinfo{journal}{\emph{arXiv preprint arXiv:1412.3555}} (\bibinfo{year}{2014}).
\newblock


\bibitem[Cong et~al\mbox{.}(2018)]%
        {cong2018xa}
\bibfield{author}{\bibinfo{person}{Qing Cong}, \bibinfo{person}{Zhiyong Feng}, \bibinfo{person}{Fang Li}, \bibinfo{person}{Yang Xiang}, \bibinfo{person}{Guozheng Rao}, {and} \bibinfo{person}{Cui Tao}.} \bibinfo{year}{2018}\natexlab{}.
\newblock \showarticletitle{XA-BiLSTM: A deep learning approach for depression detection in imbalanced data}. In \bibinfo{booktitle}{\emph{2018 IEEE international conference on bioinformatics and biomedicine (BIBM)}}. IEEE, \bibinfo{pages}{1624--1627}.
\newblock


\bibitem[Farruque et~al\mbox{.}(2021)]%
        {farruque2021explainable}
\bibfield{author}{\bibinfo{person}{Nawshad Farruque}, \bibinfo{person}{Randy Goebel}, \bibinfo{person}{Osmar~R Za{\"\i}ane}, {and} \bibinfo{person}{Sudhakar Sivapalan}.} \bibinfo{year}{2021}\natexlab{}.
\newblock \showarticletitle{Explainable zero-shot modelling of clinical depression symptoms from text}. In \bibinfo{booktitle}{\emph{2021 20th IEEE International Conference on Machine Learning and Applications (ICMLA)}}. IEEE, \bibinfo{pages}{1472--1477}.
\newblock


\bibitem[Jiang et~al\mbox{.}(2020)]%
        {jiang2020detection}
\bibfield{author}{\bibinfo{person}{Zheng~Ping Jiang}, \bibinfo{person}{Sarah~Ita Levitan}, \bibinfo{person}{Jonathan Zomick}, {and} \bibinfo{person}{Julia Hirschberg}.} \bibinfo{year}{2020}\natexlab{}.
\newblock \showarticletitle{Detection of mental health from reddit via deep contextualized representations}. In \bibinfo{booktitle}{\emph{Proceedings of the 11th International Workshop on Health Text Mining and Information Analysis}}. \bibinfo{pages}{147--156}.
\newblock


\bibitem[Kim et~al\mbox{.}(2020)]%
        {kim2020deep}
\bibfield{author}{\bibinfo{person}{Jina Kim}, \bibinfo{person}{Jieon Lee}, \bibinfo{person}{Eunil Park}, {and} \bibinfo{person}{Jinyoung Han}.} \bibinfo{year}{2020}\natexlab{}.
\newblock \showarticletitle{A deep learning model for detecting mental illness from user content on social media}.
\newblock \bibinfo{journal}{\emph{Scientific reports}} \bibinfo{volume}{10}, \bibinfo{number}{1} (\bibinfo{year}{2020}), \bibinfo{pages}{1--6}.
\newblock


\bibitem[Kim et~al\mbox{.}(2006)]%
        {kim2006pathological}
\bibfield{author}{\bibinfo{person}{Suck~Won Kim}, \bibinfo{person}{Jon~E Grant}, \bibinfo{person}{Elke~D Eckert}, \bibinfo{person}{Patricia~L Faris}, {and} \bibinfo{person}{Boyd~K Hartman}.} \bibinfo{year}{2006}\natexlab{}.
\newblock \showarticletitle{Pathological gambling and mood disorders: Clinical associations and treatment implications}.
\newblock \bibinfo{journal}{\emph{Journal of affective disorders}} \bibinfo{volume}{92}, \bibinfo{number}{1} (\bibinfo{year}{2006}), \bibinfo{pages}{109--116}.
\newblock


\bibitem[Kim and Vossen(2021)]%
        {kim2021emoberta}
\bibfield{author}{\bibinfo{person}{Taewoon Kim} {and} \bibinfo{person}{Piek Vossen}.} \bibinfo{year}{2021}\natexlab{}.
\newblock \bibinfo{title}{EmoBERTa: Speaker-Aware Emotion Recognition in Conversation with RoBERTa}.
\newblock
\newblock
\showeprint[arxiv]{2108.12009}~[cs.CL]


\bibitem[Koob(2010)]%
        {koob2010encyclopedia}
\bibfield{author}{\bibinfo{person}{George~F Koob}.} \bibinfo{year}{2010}\natexlab{}.
\newblock \bibinfo{booktitle}{\emph{Encyclopedia of behavioral neuroscience}}.
\newblock \bibinfo{publisher}{Elsevier}.
\newblock


\bibitem[Lopes(2021)]%
        {lopes2021cedri}
\bibfield{author}{\bibinfo{person}{Rui~Pedro Lopes}.} \bibinfo{year}{2021}\natexlab{}.
\newblock \showarticletitle{CeDRI at eRisk 2021: A naive approach to early detection of psychological disorders in social media}. In \bibinfo{booktitle}{\emph{CEUR Workshop Proceedings}}. CEUR Workshop Proceedings, \bibinfo{pages}{981--991}.
\newblock


\bibitem[Loyola et~al\mbox{.}(2022)]%
        {loyola2022unsl}
\bibfield{author}{\bibinfo{person}{Juan~Mart{\'\i}n Loyola}, \bibinfo{person}{Horacio Thompson}, \bibinfo{person}{Sergio Burdisso}, {and} \bibinfo{person}{Marcelo Errecalde}.} \bibinfo{year}{2022}\natexlab{}.
\newblock \showarticletitle{UNSL at eRisk 2022: Decision policies with history for early classification}.
\newblock  (\bibinfo{year}{2022}).
\newblock


\bibitem[M{\'a}rmol-Romero et~al\mbox{.}(2022)]%
        {marmol2022sinai}
\bibfield{author}{\bibinfo{person}{Alba~Mar{\'\i}a M{\'a}rmol-Romero}, \bibinfo{person}{Salud~Mar{\'\i}a Jim{\'e}nez-Zafra}, \bibinfo{person}{Flor~Miriam Plaza-del Arco}, \bibinfo{person}{M~Dolores Molina-Gonz{\'a}lez}, \bibinfo{person}{Mar{\'\i}a-Teresa Mart{\'\i}n-Valdivia}, {and} \bibinfo{person}{Arturo Montejo-R{\'a}ez}.} \bibinfo{year}{2022}\natexlab{}.
\newblock \showarticletitle{SINAI at eRisk@ CLEF 2022: Approaching Early Detection of Gambling and Eating Disorders with Natural Language Processing}.
\newblock  (\bibinfo{year}{2022}).
\newblock


\bibitem[Miller(2013)]%
        {miller2013interventions}
\bibfield{author}{\bibinfo{person}{Peter~M Miller}.} \bibinfo{year}{2013}\natexlab{}.
\newblock \bibinfo{booktitle}{\emph{Interventions for Addiction: Comprehensive Addictive Behaviors and Disorders, Volume 3}}. Vol.~\bibinfo{volume}{3}.
\newblock \bibinfo{publisher}{Academic Press}.
\newblock


\bibitem[Morales et~al\mbox{.}(2021)]%
        {morales2021team}
\bibfield{author}{\bibinfo{person}{Michelle Morales}, \bibinfo{person}{Prajjalita Dey}, {and} \bibinfo{person}{Kriti Kohli}.} \bibinfo{year}{2021}\natexlab{}.
\newblock \showarticletitle{Team 9: A Comparison of Simple vs. Complex Models for Suicide Risk Assessment}. In \bibinfo{booktitle}{\emph{Proceedings of the seventh workshop on computational linguistics and clinical psychology: Improving access}}. \bibinfo{pages}{99--102}.
\newblock


\bibitem[Murarka et~al\mbox{.}(2021)]%
        {murarka2021classification}
\bibfield{author}{\bibinfo{person}{Ankit Murarka}, \bibinfo{person}{Balaji Radhakrishnan}, {and} \bibinfo{person}{Sushma Ravichandran}.} \bibinfo{year}{2021}\natexlab{}.
\newblock \showarticletitle{Classification of mental illnesses on social media using RoBERTa}. In \bibinfo{booktitle}{\emph{Proceedings of the 12th International Workshop on Health Text Mining and Information Analysis}}. \bibinfo{pages}{59--68}.
\newblock


\bibitem[Naseem et~al\mbox{.}(2022)]%
        {naseem2022early}
\bibfield{author}{\bibinfo{person}{Usman Naseem}, \bibinfo{person}{Adam~G Dunn}, \bibinfo{person}{Jinman Kim}, {and} \bibinfo{person}{Matloob Khushi}.} \bibinfo{year}{2022}\natexlab{}.
\newblock \showarticletitle{Early identification of depression severity levels on reddit using ordinal classification}. In \bibinfo{booktitle}{\emph{Proceedings of the ACM Web Conference 2022}}. \bibinfo{pages}{2563--2572}.
\newblock


\bibitem[Parapar et~al\mbox{.}(2021)]%
        {parapar2021overview}
\bibfield{author}{\bibinfo{person}{Javier Parapar}, \bibinfo{person}{Patricia Mart{\'\i}n-Rodilla}, \bibinfo{person}{David~E Losada}, {and} \bibinfo{person}{Fabio Crestani}.} \bibinfo{year}{2021}\natexlab{}.
\newblock \showarticletitle{Overview of eRisk at CLEF 2021: Early Risk Prediction on the Internet (Extended Overview).}
\newblock \bibinfo{journal}{\emph{CLEF (Working Notes)}} (\bibinfo{year}{2021}), \bibinfo{pages}{864--887}.
\newblock


\bibitem[Parapar et~al\mbox{.}(2022)]%
        {parapar2022overview}
\bibfield{author}{\bibinfo{person}{Javier Parapar}, \bibinfo{person}{Patricia Mart{\'\i}n-Rodilla}, \bibinfo{person}{David~E Losada}, {and} \bibinfo{person}{Fabio Crestani}.} \bibinfo{year}{2022}\natexlab{}.
\newblock \showarticletitle{Overview of erisk 2022: Early risk prediction on the internet}. In \bibinfo{booktitle}{\emph{Experimental IR Meets Multilinguality, Multimodality, and Interaction: 13th International Conference of the CLEF Association, CLEF 2022, Bologna, Italy, September 5--8, 2022, Proceedings}}. Springer, \bibinfo{pages}{233--256}.
\newblock


\bibitem[Ragheb et~al\mbox{.}(2021)]%
        {ragheb2021negatively}
\bibfield{author}{\bibinfo{person}{Waleed Ragheb}, \bibinfo{person}{J{\'e}r{\^o}me Az{\'e}}, \bibinfo{person}{Sandra Bringay}, {and} \bibinfo{person}{Maximilien Servajean}.} \bibinfo{year}{2021}\natexlab{}.
\newblock \showarticletitle{Negatively Correlated Noisy Learners for At-Risk User Detection on Social Networks: A Study on Depression, Anorexia, Self-Harm, and Suicide}.
\newblock \bibinfo{journal}{\emph{IEEE Transactions on Knowledge and Data Engineering}} \bibinfo{volume}{35}, \bibinfo{number}{1} (\bibinfo{year}{2021}), \bibinfo{pages}{770--783}.
\newblock


\bibitem[Rao et~al\mbox{.}(2020)]%
        {rao2020knowledge}
\bibfield{author}{\bibinfo{person}{Guozheng Rao}, \bibinfo{person}{Chengxia Peng}, \bibinfo{person}{Li Zhang}, \bibinfo{person}{Xin Wang}, {and} \bibinfo{person}{Zhiyong Feng}.} \bibinfo{year}{2020}\natexlab{}.
\newblock \showarticletitle{A knowledge enhanced ensemble learning model for mental disorder detection on social media}. In \bibinfo{booktitle}{\emph{Knowledge Science, Engineering and Management: 13th International Conference, KSEM 2020, Hangzhou, China, August 28--30, 2020, Proceedings, Part II 13}}. Springer, \bibinfo{pages}{181--192}.
\newblock


\bibitem[Sawhney et~al\mbox{.}(2020)]%
        {sawhney2020time}
\bibfield{author}{\bibinfo{person}{Ramit Sawhney}, \bibinfo{person}{Harshit Joshi}, \bibinfo{person}{Saumya Gandhi}, {and} \bibinfo{person}{Rajiv Shah}.} \bibinfo{year}{2020}\natexlab{}.
\newblock \showarticletitle{A time-aware transformer based model for suicide ideation detection on social media}. In \bibinfo{booktitle}{\emph{Proceedings of the 2020 conference on empirical methods in natural language processing (EMNLP)}}. \bibinfo{pages}{7685--7697}.
\newblock


\bibitem[Sophia and Zilberman(2013)]%
        {SOPHIA20131305}
\bibfield{author}{\bibinfo{person}{Eglacy~C. Sophia} {and} \bibinfo{person}{Monica~L. Zilberman}.} \bibinfo{year}{2013}\natexlab{}.
\newblock \showarticletitle{Chapter 88 - Addictive Disorders}.
\newblock In \bibinfo{booktitle}{\emph{Women and Health (Second Edition)} (\bibinfo{edition}{second edition} ed.)}, \bibfield{editor}{\bibinfo{person}{Marlene~B. Goldman}, \bibinfo{person}{Rebecca Troisi}, {and} \bibinfo{person}{Kathryn~M. Rexrode}} (Eds.). \bibinfo{publisher}{Academic Press}, \bibinfo{pages}{1305--1316}.
\newblock
\showISBNx{978-0-12-384978-6}
\urldef\tempurl%
\url{https://doi.org/10.1016/B978-0-12-384978-6.00088-1}
\showDOI{\tempurl}


\bibitem[Tadesse et~al\mbox{.}(2019)]%
        {tadesse2019detection}
\bibfield{author}{\bibinfo{person}{Michael~Mesfin Tadesse}, \bibinfo{person}{Hongfei Lin}, \bibinfo{person}{Bo Xu}, {and} \bibinfo{person}{Liang Yang}.} \bibinfo{year}{2019}\natexlab{}.
\newblock \showarticletitle{Detection of suicide ideation in social media forums using deep learning}.
\newblock \bibinfo{journal}{\emph{Algorithms}} \bibinfo{volume}{13}, \bibinfo{number}{1} (\bibinfo{year}{2019}), \bibinfo{pages}{7}.
\newblock


\bibitem[Turcan et~al\mbox{.}(2021)]%
        {turcan2021emotion}
\bibfield{author}{\bibinfo{person}{Elsbeth Turcan}, \bibinfo{person}{Smaranda Muresan}, {and} \bibinfo{person}{Kathleen McKeown}.} \bibinfo{year}{2021}\natexlab{}.
\newblock \showarticletitle{Emotion-infused models for explainable psychological stress detection}. In \bibinfo{booktitle}{\emph{Proceedings of the 2021 conference of the North American Chapter of the Association for Computational Linguistics: human language technologies}}. \bibinfo{pages}{2895--2909}.
\newblock


\bibitem[Uban et~al\mbox{.}(2021)]%
        {uban2021explainability}
\bibfield{author}{\bibinfo{person}{Ana~Sabina Uban}, \bibinfo{person}{Berta Chulvi}, {and} \bibinfo{person}{Paolo Rosso}.} \bibinfo{year}{2021}\natexlab{}.
\newblock \showarticletitle{On the explainability of automatic predictions of mental disorders from social media data}. In \bibinfo{booktitle}{\emph{Natural Language Processing and Information Systems: 26th International Conference on Applications of Natural Language to Information Systems, NLDB 2021, Saarbr{\"u}cken, Germany, June 23--25, 2021, Proceedings}}. Springer, \bibinfo{pages}{301--314}.
\newblock


\bibitem[Vaswani et~al\mbox{.}(2017)]%
        {vaswani2017attention}
\bibfield{author}{\bibinfo{person}{Ashish Vaswani}, \bibinfo{person}{Noam Shazeer}, \bibinfo{person}{Niki Parmar}, \bibinfo{person}{Jakob Uszkoreit}, \bibinfo{person}{Llion Jones}, \bibinfo{person}{Aidan~N Gomez}, \bibinfo{person}{{\L}ukasz Kaiser}, {and} \bibinfo{person}{Illia Polosukhin}.} \bibinfo{year}{2017}\natexlab{}.
\newblock \showarticletitle{Attention is all you need}.
\newblock \bibinfo{journal}{\emph{Advances in neural information processing systems}}  \bibinfo{volume}{30} (\bibinfo{year}{2017}).
\newblock


\bibitem[Wang et~al\mbox{.}(2021)]%
        {wang2021learning}
\bibfield{author}{\bibinfo{person}{Ning Wang}, \bibinfo{person}{Fan Luo}, \bibinfo{person}{Yuvraj Shivtare}, \bibinfo{person}{Varsha~D Badal}, \bibinfo{person}{KP Subbalakshmi}, \bibinfo{person}{Rajarathnam Chandramouli}, {and} \bibinfo{person}{Ellen Lee}.} \bibinfo{year}{2021}\natexlab{}.
\newblock \showarticletitle{Learning models for suicide prediction from social media posts}.
\newblock \bibinfo{journal}{\emph{arXiv preprint arXiv:2105.03315}} (\bibinfo{year}{2021}).
\newblock


\bibitem[Wu et~al\mbox{.}(2020)]%
        {wu2020deep}
\bibfield{author}{\bibinfo{person}{Min~Yen Wu}, \bibinfo{person}{Chih-Ya Shen}, \bibinfo{person}{En~Tzu Wang}, {and} \bibinfo{person}{Arbee~LP Chen}.} \bibinfo{year}{2020}\natexlab{}.
\newblock \showarticletitle{A deep architecture for depression detection using posting, behavior, and living environment data}.
\newblock \bibinfo{journal}{\emph{Journal of Intelligent Information Systems}}  \bibinfo{volume}{54} (\bibinfo{year}{2020}), \bibinfo{pages}{225--244}.
\newblock


\bibitem[Zogan et~al\mbox{.}(2022)]%
        {zogan2022explainable}
\bibfield{author}{\bibinfo{person}{Hamad Zogan}, \bibinfo{person}{Imran Razzak}, \bibinfo{person}{Xianzhi Wang}, \bibinfo{person}{Shoaib Jameel}, {and} \bibinfo{person}{Guandong Xu}.} \bibinfo{year}{2022}\natexlab{}.
\newblock \showarticletitle{Explainable depression detection with multi-aspect features using a hybrid deep learning model on social media}.
\newblock \bibinfo{journal}{\emph{World Wide Web}} \bibinfo{volume}{25}, \bibinfo{number}{1} (\bibinfo{year}{2022}), \bibinfo{pages}{281--304}.
\newblock


\end{thebibliography}

\appendix

\end{document}